# FedWOA: A Federated Learning Model that uses the Whale Optimization Algorithm for Renewable Energy Prediction

Viorica Chifu, Tudor Cioara, Cristian Anitiei, Cristina Pop, Ionut Anghel

Computer Science Department, Technical University of Cluj-Napoca, Memorandumului 28,

400114 Cluj-Napoca, Romania

{viorica.chifu; tudor.cioara; cristina.pop; ionut.anghel}@cs.utcluj.ro, anitei.va.cristian@student.utcluj.ro

**Abstract:** Privacy is important when dealing with sensitive personal information in machine learning models, which require large data sets for training. In the energy field, access to household's prosumer energy data is crucial for energy predictions to support energy grid management and large-scale adoption of renewables however citizens are often hesitant to grant access to cloud-based machine learning models. Federated learning has been proposed as a solution to privacy challenges however report issues in generating the global prediction model due to data heterogeneity, variations in generation patterns, and the high number of parameters leading to even lower prediction accuracy. This paper addresses these challenges by introducing FedWOA a novel federated learning model that employs the Whale Optimization Algorithm to aggregate global prediction models from the weights of local LTSM neural network models trained on prosumer energy data. The proposed solution identifies the optimal vector of weights in the search spaces of the local models to construct the global shared model and then is subsequently transmitted to the local nodes to improve the prediction quality at the prosumer site while for handling non-IID data K-Means was used for clustering prosumers with similar scale of energy data. The evaluation results on prosumers' energy data have shown that FedWOA can effectively enhance the accuracy of energy prediction models accuracy by 25% for MSE and 16% for MAE compared to FedAVG while demonstrating good convergence and reduced loss.

## 1. Introduction

To be effective machine learning models use huge amounts of data in the training processes. As the data may contain sensitive information about individuals ensuring privacy in learning while still obtaining good performance is crucial [1]. This is valid for many sectors where data privacy and decentralization are important such as healthcare, IoT, or energy. For example, in the energy domain, the transition towards renewable energy shifted the focus to citizens and energy prosumers who need to be actively involved in the energy management process [2]. More decentralized energy ecosystems are emerging that require the active participation of individuals and communities in managing their energy production and consumption [3]. To balance energy supply and demand and manage the local grid stability, access to households' prosumers' energy data is required for training energy prediction machine learning models to support the development of adaptation and proactive optimization strategies [4]. Thus, lately, data-driven machine learning modes have been widely proposed for energy predictions [5], [6], [7] like regression models good in analyzing historical energy and weather data, deep learning models that may identify long-term dependencies in energy data, or other machine learning models can capture complex relationships between input features and energy production. For better accuracy and effectiveness, they use a combination of features derived from energy data, timestamps, and contextual weather information. However, the households' prosumers' energy data contains sensitive information that is rather private. They need to be carefully handled and protected when used in energy management applications thus techniques like data anonymization [8], encryption [9], and differential privacy [10] have recently emerged. But even with strong privacy and security guarantees the citizen are often reluctant in giving access to their energy data to be moved in centralized silos in the cloud to be further processed and used in training by taking advantage of the cloud's potentially unlimited computing resources. Moreover, in Europe, it is crucial to comply with the General Data Protection Regulation (GDPR) when using machine learning in the cloud aspects like data minimization, retention, deletion, or cross-border transfer affecting the effectiveness of the training processes [11].

In this context, federated learning (FL) models have been proposed lately specifically for cases in which the data owners do not want to share their data due to privacy concerns [12]. They are effective in ensuring the confidentiality and security of prosumers data and they enable multiple local nodes to collaboratively train a machine learning model without sharing their data, which is kept locally [13]. The local machine models are trained on the citizen's devices and are then transmitted to a central server that aggregates and integrates them into a shared global model. The central server sends back the learned global model to the citizen's devices for further refinement and eventually usage. The FL cycle is repeated several times before the global model reaches the desired optimal accuracy. As it is a critical process for the creation of the global federated models several techniques have been proposed lately, the most representative ones being federated averaging of weights (FedAVG), Federated Stochastic Gradient Descent (FedSGD) and their combinations [14], [15]. In the FedAVG case, the aggregation of the local models learned to infer the global shared model is done by averaging (e.g., weighted, median, trimmed, etc.) their model updates in terms of weights while for the FedSGD the local nodes compute and share the gradients of the loss function concerning their local model parameters using their local data. However, even if these approaches generally provide acceptable solutions, there are situations in which the global model obtained with FL has lower accuracy than the locally learned models [16]. This may be caused by the rather simplistic approach of combining the learned weights from the local nodes. Challenges are reported in generating the global model due to local energy data heterogeneity, variations in generation patterns, and the high number of parameters that must be considered especially in the case of deep learning making the averaging not appropriate [17].

Advanced aggregation methods need to be researched and developed to deal with the complexity of the problem search spaces. In this context, we consider the bio-inspired population-based heuristics good candidates for performing more optimized aggregation at the central node level than the averaging-based solutions. The population-based metaheuristics use the initialization and iterative evolution phases to explore the solution space and produce high-quality solutions [18]. The initial population is generated and evaluated, with the best individual identified. Search agents, inspired by living organisms, are generated randomly, or using some prior knowledge. A fitness function measures everyone's quality in the search space. Iterative evolution involves updating a population using a mathematical model inspired by animal behavior. Fitness is evaluated to find the best search agent, which is then iteratively updated if a better one is found. The best individual from all populations is returned as the solution. As a result, the population-based metaheuristics guide the search process, efficiently explore the solution space, prevent local optima, use a wide range of algorithms, are not specific to certain problems, and can use domain-specific knowledge through local heuristics controlled by a higher level. Among the metaheuristics proposed in the last years, one of the most promising and that has shown effectiveness in some cases in energy engineering and optimization cases is the Whale Optimization Algorithm (WOA) [19]. It is inspired by the hunting behavior of humpback whales that balances the exploration and exploitation of the solution space by guiding the whales to search for new regions while converging towards an optimum solution. WOA ensures a good balance between exploration and exploitation, with low computational complexity and a good convergence rate in a variety of optimization use cases [20], [21]. Given these qualities, WOA emerges as a promising candidate for addressing the optimization challenges posed by federated models.

In this paper, we propose FedWOA, a novel FL model for time series data that uses the WOA to construct the global shared model by searching and identifying the optimal weights in a search space given by the learned local model vectors of weights. The generated FL model is subsequently transmitted to the local nodes to enhance the quality of local models. To show FedWOA effectiveness we have used it for predicting renewable energy generation in the case of households' energy prosumers that train Long Short-Term Memory neural network on their local energy data. To address the problem of non-IID data in renewable energy production where we need to deal with data heterogeneity and variations in energy generation patterns, we apply the K-Means clustering algorithm to group prosumers with a similar scale of energy data into clusters. Experimental results show that FedWOA achieves better accuracy in energy prediction compared to FedAVG for both Mean Square Error (MSE) and MAE metrics.

The rest of the paper is structured as follows: Section 2 presents the state-of-the-art models for federated machine learning in different domains with a focus on energy prediction cases, Section 3 describes the FL model using WOA for global model construction, Section 4 presents the relevant results in the context of prosumers energy prediction comparing the accuracy with federated average models, Section 5 discusses the convergence rate and diversity of a WOA for the FL, while Section 6 presents conclusions and future work.

## 2. Related work

There are limited FL strategies applied in the context of energy grid management, and these strategies primarily focus on areas such as privacy-preserving energy prediction, demand response, and energy optimization. Fekri et al. use stochastic gradient descent (FedSGD) and federated average (FedAVG) to predict the loads in an energy network [23]. During training, the FedSGD sends updates after each round, while the FedAVG uses multiple batches per node. A secure federated deep learning solution for heating load prediction in a building environment is proposed in [24]. To prevent overfitting and maintain confidentiality of data, a dropout regularization technique is used to transmit gradient updates to the central node. Savi et al. [25] apply FL in edge computing scenarios using Long Short-Term Memory (LSTM) models on users' devices and user-specific historical energy consumption data and then are aggregated into a global model on a central node. The novelty lies in the users' clustering based on socioeconomic and consumption similarities to form smaller federations that can be trained separately to produce specific federated models. Fernandez et al. [26] apply FL approaches in different scenarios that focus on the privacy of the data. Each local node adds noise to the data, a differential privacy technique to ensure that attackers cannot reconstruct the original data from gradients communicated over the network. A secure aggregation algorithm is introduced to encrypt the weights of the local models. FL is applied on different campuses to predict the load of combined cooling, heating, and power systems [27]. Several models have been implemented using different strategies and then are comparatively analyzed, the FedAdagrad providing the best prediction results under a Factitious disorder imposed on another (FDIA) attack. Shi et al. [28] combine FL with transfer learning to improve the accuracy of residential short-term load forecasting. FL is used for issues related to accessibility and privacy of the data, while transfer learning is used to handle the challenge of working with non-identically distributed data. The FL architecture is built on a hybrid model that combines a CNN with a bi-directional LSTM network. Gholizadeh et al. forecast both individual and aggregated electricity demand using FedAVG and LSTM on local nodes [29]. Before training the consumers are clustered using the energy consumption data similarity. Similarly, in [30], two steps are defined to forecast the individual consumer electricity load. First, the local models are trained on nodes and then aggregated into a global model on the central node. Second, the clients customize the global model by training it on their data, as the data is not independent and identically distributed. Liu et al. use a FL framework based on an improved Gate Recurrent Unit for forecasting distributed short-term individual load [31]. The framework guarantees the data privacy of consumers and the usage of computing capacity in edge devices by transmitting the forecasting models during model training instead of the actual load data. BuildFL platform [34] is used to predict the distributed energy resources demand and consumption [32]. The server node stores the parameters of the global model and sends it to the client nodes to train the local models using their datasets and update the model parameters. The updates are sent back to the global model so that consumer data privacy is preserved. A side channel analysis to address the vulnerability of load prediction to attack methods is proposed in [33]. The side-channel analysis extracts power-related data from a chip that executes a secure FL load forecasting model, and an enhanced convolutional neural network is employed to predict the load using the extracted side-channel data. Finally, Dogra et al. [35] combine time-series pattern analysis with clustering and FL to preserve consumer data privacy, to improve load prediction accuracy. Affinity clustering groups consumers based on consumption patterns, while FL estimates energy load and improves communication efficiency.

FL techniques have been used primarily in fields, such as healthcare, autonomous driving, IoT, and edge computing, demonstrating good potential to share knowledge and insights among different local parties while simultaneously preserving the privacy of sensitive data. Brisimi et al. [36] integrate a distributed sparse Support Vector Machine (SVM) approach that works well with a small number of features extracted from EHR data and Primal-Dual Splitting clustering to differentiate between patients who are likely to be hospitalized. Similarly, in [37] sequential and Empirical Bayes based Hierarchical Bayesian federated models are proposed for heart rate prediction. Brophy et al. [38] define a FL-based method for estimating continuous blood pressure from optical photoplethysmogram signals. The global model demonstrates effectiveness for novel data and is responsible for updating the client nodes with the accumulated global weights. A similar FL architecture is proposed in [39] for referable diabetic retinopathy classification, while Chen et al. [44] propose a federal transfer-based learning framework for classifying human activities using CNN-based architecture transfer learning for model customization. Various federated models are proposed for COVID-19 classification [42] using Generative Adversarial Network and stochastic gradient descent optimizer, detection of lung lesions after COVID-19 from CT images using CNN and federated average [41], identification of COVID-19 pneumonia [40] and mortality using MLP [43]. FL has been applied to predict the driver's behavior, the vehicle trajectory, and traffic flow or to detect objects in traffic, such as pedestrians, traffic signs, or other cars, to reduce the risk of accidents and increase people's safety. Various model has been proposed such as FL with holomorphic encryption for improved security [47], spike neural networks-based architecture or one-class support vector machine with federated average aggregation [49], [50] or Bayesian convolutional neural with an uncertainty-weighted asynchronous aggregation [45]. Hossain et al. [46] store the local model updates on blockchain to improve global model traceability and use a differential privacy method with a noise-adding mechanism to improve the confidentiality of the model against attacks of data poisoning. In the IoT and edge computing domains, which are relevant to smart grid management, FL is employed for purposes such as anomaly and intrusion detection, as well as data privacy protection. Wu et al. [51] propose a personalized FL-based framework for IoT applications to manage heterogeneous data while maintaining data privacy. Federation average is used to learn a global shared model based on the local model updates provided by the client nodes and then is customized based on personal data. Lazzarini et al. [54] use FL for IoT intrusion detection. The client nodes train local models using a shallow artificial neural network, and the server node aggregates local models using the federated average. In the context of resource management or network communication, FL-based solutions address problems such as malicious node detection, unreliable communication, and bandwidth limitation. Fu et al. [52] use blockchain in FL to ensure security for exchanging the local model parameters. Yang et al. [57] have put forward three scheduling policies to evaluate the performance of FL in a resource-limited setting with restricted bandwidth and transmission interference, while the problem of ultra-reliable low-latency communication in vehicular networks is addressed in [56] by combining FL with Lyapunov Optimization Algorithm to reduce the delays.

We have identified only a few state-of-the-art FL solutions that focus on global model optimization using heuristics, but neither of these solutions has been applied in the context of smart grid management. This highlights the current gap or lack of specific solutions in the smart grid management domain on using heuristics for FL optimization. Thavavel et al. [55] combine FL with metaheuristic algorithms to perform intrusion detection in IoT environments. The local nodes integrate a social group optimization algorithm kernel extreme learning machine for model training, and the bird swarm algorithm is used to select the most relevant features for training the algorithms. A combination of FL and particle swarm optimization was proposed in [53] to enhance network communication and decrease the amount of data being transmitted. Unlike traditional methods where the weights of the models learned on local nodes are sent to the server, this approach uses the PSO algorithm to deduce the score values (i.e., accuracy or loss) of the learned models. Finally, a method for detecting anomalies in pedestrians from remote sensing data was proposed in [48], combining FL with Harris Hawks Optimizer. FL is integrated with a deep faith network architecture to perform anomaly detection, while the Harris Hawks Optimizer is employed to adjust network hyperparameters.

In the case of FL approaches for smart grid applications, challenges are reported in generating the global model due to local energy data heterogeneity, and variations in generation patterns. Consequently, when such diverse local machine learning models are combined into a single global model, decreases in energy prediction accuracy can be noticed. We have addressed this by adapting and using the WOA to construct the global shared model by identifying the optimal vector of weights, which is subsequently transmitted to the local nodes to enhance the quality of local models. The issue of non-IID data was addressed by initially performing clustering of the prosumers to group local nodes with similar scale of energy data. The experimental results show that our FL combined with the WOA yields better results compared with FedAVG algorithm when applied to the prosumer energy generation prediction problem. Also, FedWOA can efficiently explore the search space and identify a solution close to the global optimum achieving a better loss (fitness) in just ten iterations, whereas FedAVG requires 50 epochs to achieve a similar result.

## 3. FL model

The FL architecture is presented in Figure 1. It consists of multiple local nodes geographically distributed. Each node acquires data using IoT metering devices in the form of time-series and stores it locally. Each cluster groups a set of local nodes that have homogeneous data and has a central node responsible for node coordination and federated model aggregation.

### 3.1. Model assumptions

We have considered that the data used in the learning process is distributed across several local nodes and each node is owned by a different user or organization. The data is geographically distributed and stored in $K$ local nodes and not shared with the central server or other participants, respecting the privacy-preserving requirements while allowing for better utilization of resources, easier maintenance, and avoiding a single point of failure:

$$d_i = \{data \mid stored\ by\ n_i\ under\ privacy\ policy\ \vartheta_i\} \tag{1}$$

The data explored in this model is in the form of a time series, the data points being ordered and indexed by time, $t$. It represents a stream of observations, $o$, recorded at specific and equal time intervals by IoT measuring devices:

$$data = \{(t, o): t \in T\ \wedge\ o \in \mathbb{R}\} \tag{2}$$

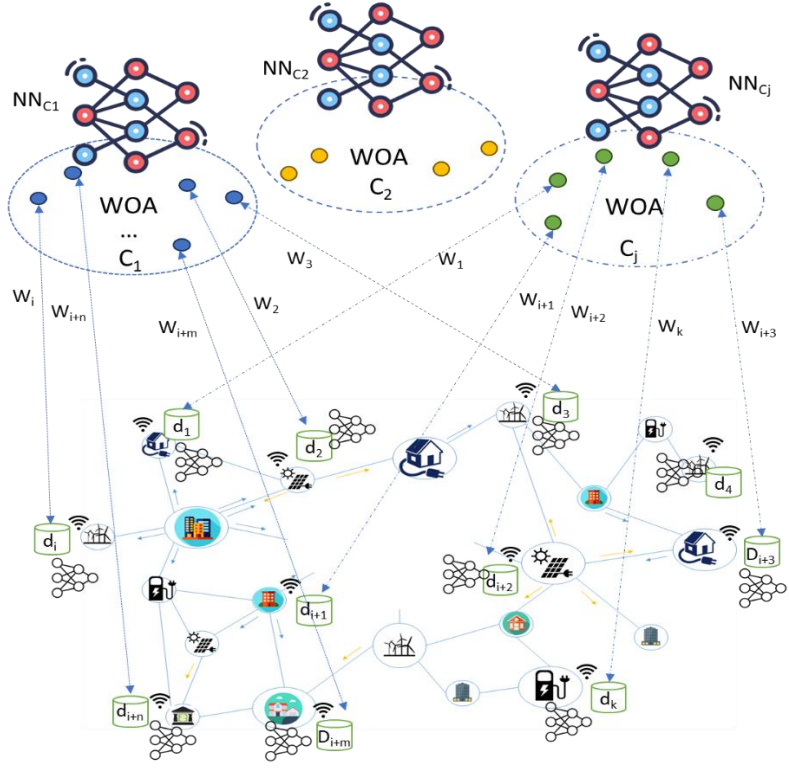

*Figure 1: FL architecture*

On each local node data $d_i$ a neural network ($NN$) is trained and only the local model parameters are shared among the nodes to follow the privacy and data security policy $\vartheta_i$:

$$NN \xrightarrow{train\ with\ d_i} NN_i \tag{3}$$

We assume that in each node the same neural network architecture is used in training, and $W_i$ is the set of weight vectors specific to the trained model in the local node $i$:

$$W_i = [w_{i,1}, \ldots, w_{i,j}, \ldots w_{i,M}], W_i \in NN_i \tag{4}$$

where $w_{i,j} \in \mathbb{R}$ is a model parameter and $M$ is the maximum number of parameters of the local model $NN_i$. The objective of the training process at the level of each local node $n_i$ is to identify a weight vector $W_i$ that minimize a loss function for the $P$ steps in the future:

$$L_n: M \times K \to \mathbb{R}, L_{n_i}(W_i) = \frac{1}{p}\sum_{p=1}^{P} MSE(E_p, O_p) \tag{5}$$

where MSE is Mean Squared Error, $E_p$ is the estimated value by the local model and $O_p$ are the actual value provided by the IoT device.

In the case of smart grid management scenarios, it is common to have data in the local nodes that can differ in characteristics and distributions due to different user preferences, device types, or user behaviour. To manage this, we have considered the use of clustering algorithms to group the nodes that have data with similar statistical characteristics or distributions before the training of the local ML model:

$$C = \{n_c, c \in 1..k \wedge f_d(n_c, n_{c+1}) < \varepsilon\} \tag{6}$$

where $f_d$ is a function that measures the similarity among data using various criteria such as maxim, minim, medium values, pattern matching, etc. Also, we considered the clusters to be distinct and there is no overlap among them:

$$\forall\ C_i, C_j \in \mathbb{C}, C_i \cap C_j = \emptyset \tag{7}$$

At the level of each cluster, we aim to infer a globally distributed model by optimally aggregating each local model weight to benefit from knowledge acquired from each local node data:

$$NN_C = AGG\ (W_i), \forall\ n_i \in C, W_i\ paramters\ of\ local\ model\ NN_i \tag{8}$$

In each cluster, we have a central node $N_c$ that runs a metaheuristic algorithm to identify the best vector of weights (i.e., model parameters) that produces the best global model for the cluster obtaining the minimum global loss:

$$GL = \min_{heuristic} L_{N_C}(NN_C) \tag{9}$$

The optimization in this case involves searching in the solutions' search space for an optimal solution that minimizes a set of given fitness functions under certain constraints [xx]:

$$(\omega_C, L, \Omega) \tag{10}$$

where $\omega_C$ is the set of candidate solutions (i.e., vector parameters of the cluster model), $L$ is the loss function or the objective functions that need to be minimized, and $\Omega$ is a set of constraints that must be fulfilled. The domain of values of the decision variables forms the search space of the optimization problem. The set of constraints, $\Omega$, is formally defined for the learned model parameters as:

$$\Omega = \{W_{NN_C} \geq 0\} \tag{11}$$

The weights aggregation process and cluster model construction are optimal in preserving overall accuracy and performance. After FL is performed at the level of each cluster the local model weight of the nodes is updated if it yields a better performance:

$$L_{N_C}(NN_C) \geq L_{n_i}(W_i), \forall\ n_i \in C \tag{12}$$

### 3.2. WOA for distributed learning

The heuristic starts with an initial population of candidate solutions and iteratively updates the population by reproducing the three phases of whale behaviour: search and circling the prey, and bubble-net attacking. These steps are used to update the position of each solution within the population.

In our federated case, the space of candidate solutions is represented by vectors of parameters associated with the neural network model:

$$\omega_C: MxM, \omega_C = \{\ \vec{W_i}\ |\ \vec{W_i} = [w_{i,1}, \ldots, w_M], w_{i,j} \in \mathbb{R}\} \tag{13}$$

Every individual has a position in the candidate solution space given by the actual values of the associated vector of weights.

The initial population of individuals is generated based on the weight vectors corresponding to the local nodes where models were trained, ensuring a good representation of each local node. Also, it ensures that the optimization algorithm uses a population of weight vectors already trained using local data. The local nodes perform gradient updates to change the local models' weights during training. The modified weight vectors are returned by each local node and are used to create the initial population for the WOA. Each candidate solution (i.e., weight vector in the population) is evaluated using the defined loss function, and the fitness values (GL) are used to guide its search for the best solution.

The Bubble-net attacking phase has two main steps: shrinking encircling and spiral updating position. In the shrinking encircling step, everyone within the population tries to improve their position relative to the best individual. This means that each member of the population tries to adjust their position ($\overrightarrow{W_i'}$) to approximate the location occupied by the best individual within the population:

$$\overrightarrow{W_i'} = \overrightarrow{W_{best}} - \vec{A} * \vec{D} \tag{14}$$

where $A$ and $D$ are vectors of the same length as $\overrightarrow{W'}_i$, while $\overrightarrow{W_{best}}$ is the vector of weights corresponding to the best individual $i$ at the current iteration. We assess the loss of all individuals based on their weight vector to determine the best individual. The two vectors $\vec{A}$ and $\vec{D}$ and are defined as follows:

$$\vec{D} = |\vec{C} * \overrightarrow{W_{best}} - \overrightarrow{W_i}| \tag{15}$$

$$\vec{A} = 2 * a * \vec{r} - a \tag{16}$$

$$\vec{C} = 2 * \vec{r} \tag{17}$$

where $\overrightarrow{W_{best}}$ is the vector of weights that corresponds to the best individual in the current iteration and $\overrightarrow{W_i}$ is the vector of weights that corresponds to the individual $i$ in the current iteration, $\vec{r}$ is a vector of random values generated in the [0,1] with the same length as $\overrightarrow{W_i}$, and $a$ is a scalar value that linearly decreases in each iteration from an initial value to zero.

In the spiral position update step, individuals in the population update their position based on their current position and the position of the best individual in the population:

$$\overrightarrow{W_i'} = \vec{D}' * e^{bl} * \cos(2\pi l) + \overrightarrow{W_{best}} \tag{18}$$

where, $b$ is a constant, $l$ is a random value in $[-1,1]$, and $\vec{D}'$ is a weights vector defined as:

$$\vec{D}' = \overrightarrow{W_{best}} - \overrightarrow{W_i} \tag{19}$$

The Bubble-net attacking phase combines the encircling and spiral updating position steps using the formula below:

$$\overrightarrow{W_i'} = \begin{cases} \overrightarrow{W_{best}} - \vec{A} * \vec{D}, & if\, p < 0.5 \\ \vec{D}' * e^{bl} * \cos(2\pi l) + \overrightarrow{W_{best}}, & otherwise \end{cases} \tag{20}$$

where $p$ is a random number in [0,1].

A subset of whales will move towards the current best solution in the population, as in the original WOA, while for the remaining whales (those not participating in the encircling prey phase) a modified update strategy was defined. Instead of adjusting the position of the current search individual towards another randomly chosen one (which might cause heavy loss drift), we propose to update the weights vector of an individual by training it on its local data using gradient updates:

$$\overrightarrow{W_i'} = \overrightarrow{W_i} - \eta * \nabla L(\overrightarrow{W_i}) \tag{21}$$

$\overrightarrow{W_i}$ is the weights vector in the current iteration, $\overrightarrow{W_i'}$ is the updated weights vector of the individual for the next iteration, $\eta$ is the learning rate of the local model, and $\nabla L(\vec{W})$ is the gradient of the loss function concerning the weights during local training.

The updated strategy has a regularizing effect. Each agent during search space exploration will try to pull the federated solution in the direction of its local optimum, thus learning more from the local data. This exploration will be valid only in the first half of the optimization process, as given by the value of $A$ that decreases from 2 to 0 across a fixed number of iterations. Thus, only when the random number $p$ is less than 0.5 and the magnitude of the vector $A$ is greater than or equal to 1, the local training will be used. These conditions will ensure a balance between exploration and exploitation of the existing solutions. After the exploration and local search phases, the positions of all whales are updated based on the solutions obtained from both the encircling prey and the local gradient-based update.

Figure 2 shows the pseudocode for the adapted WOA for FL on time series data consisting of initialization and iterative search phases. During the initialization phase, an initial population of a set of weight vectors is built by training models on local nodes using local data (see lines 2-5).

---

**Algorithm 1**

**Inputs:** $n_i$ is the set of local nodes from a cluster $C$, $d_i$ represents the data stored by each local node and used for local model training, $T$ is the maximum number of iterations, $b$ is a constant value used to define the logarithmic spiral shape

**Outputs:** $\overrightarrow{W_{best}}$ — the best vector of weights corresponding to the optimal FL model for cluster $C$

**Begin**
1. Population $= \emptyset$, Fitness $= \emptyset$, $\vec{A} = \emptyset$, $\vec{C} = \emptyset$
2. **Foreach** $n_i$ in $C$ **do**
3.    $\overrightarrow{W_i} = TRAIN\_ON\_NODE\ (NN_i, d_i)$
4.    $UPDATE\_POPULATION(\overrightarrow{W_i})$
5. **End for**
6. **Foreach** $\overrightarrow{W_i}$ in Population **do**
7.    $L_{n_i} = COMPUTE\_LOCAL\_LOSS(n_i, \overrightarrow{W_i}, d_i)$
8. **End for**
9. $GL = COMPUTE\_GLOBAL\_LOSS(L_{n_i}, \overrightarrow{W_i}, |C|)$
10. $\overrightarrow{W_{best}} = SELECT\_MINIMUM\_LOSS(Population, GL)$
11. $t = 0, a = 2$
12. **while** (t < T) **do**
13.   **foreach** $\overrightarrow{W_i}$ in Population **do**
14.     $LINIAR\_DECREASE(a)$
15.     $p = RANDOM\_NUMBER(0,1); l = RANDOM\_NUMBER(-1,1)$
16.     $\vec{r} = RANDOM\_VECTOR(0,1)$
17.     $\vec{A} = UPDATE(\vec{r}, a); \vec{C} = UPDATE(\vec{r})$
18.     if (p < 0.5)
19.       if (|A| < 1) $\overrightarrow{W_i} = UPDATE(\overrightarrow{W_i}, \overrightarrow{W_{best}}, \vec{A}, \vec{C})$
20.       else $\overrightarrow{ws_i} = TRAIN\_ON\_NODE\ (NN_i, d_i)$
21.     else $\overrightarrow{W_i} = UPDATE(\overrightarrow{W_i}, \overrightarrow{W_{best}}, b, l)$
22.   **end for**
23.   **Foreach** $\overrightarrow{W_i}$ in Population **do**
24.     $L_{n_i} = COMPUTE\_LOCAL\_LOSS(n_i, \overrightarrow{W_i}, d_i)$
25.   **End for**
26.   $GL_{n_i} = COMPUTE\_GLOBAL\_LOSS(L_{n_i}, \overrightarrow{W_i}, |C|)$
27.   $\overrightarrow{W_{best}} = SELECT\_MINIMUM\_LOSS(Population, GL)$
28.   $t = t + 1$
29. **end while**
30. return $\overrightarrow{W_{best}}$
**End**

*Figure 2: Federated model optimization using WOA.*

The global loss is calculated for each weight vector in the population using the local loss values. From the set of all weight vectors in the population, the algorithm chooses the weight vector with the lowest overall loss value that represents the best individual (see lines 6-10). In the iterative phase, the algorithm updates the population of weight vectors by applying exploration and exploitation strategies to each vector of weights in the population (see lines 13-22). These strategies aim to explore new solutions and exploit the most promising ones. At the end of each iteration, the algorithm identifies the best weight vector within the updated population by evaluating the global loss values of the weight vectors and selecting the one with the lowest loss as the best vector of weights (see lines 23-27).

## 4. Evaluation results

To assess the effectiveness of the FL solution, we utilized a dataset comprising data from more than 30 small-scale prosumers who have their own photovoltaic (PV) system for energy production [58]. The FL model was used for renewable energy prediction enabling multiple local prosumer nodes to collaboratively train a global energy prediction model while keeping their data decentralized. The energy production profiles corresponding to each of the monitored days in the data set show how much energy is produced in each hour. Moreover, for every hour we have energy values collected every 15 minutes, to capture energy measurements at a higher level of granularity.

The K-means algorithm was used to group the local nodes that have similar statistical properties or distributions in their data subsets. Because the clustering algorithm was applied on *n*-dimensional feature vectors (in our case *n* = 7), to visually represent the formed clusters in a two-dimensional space we have applied the Principal Components Analysis (PCA) as a dimensionality reduction technique. We apply PCA on the dataset to identify and retain the principal components that capture the maximum variance in the data and use them to transform your original data points into this new two-dimensional space. After applying the clustering algorithm, we obtained 3 clusters a cluster with 9 local nodes, one with 10 local nodes, and another one with 12 local nodes (see Figure 3).

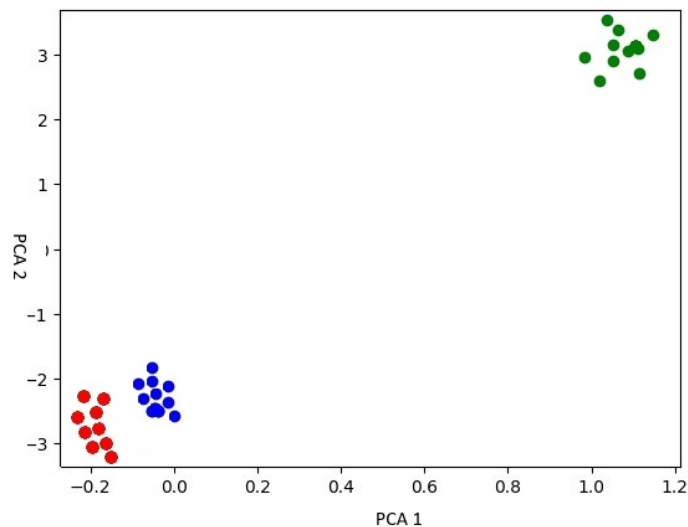

*Figure 3: Prosumers cluster visualization using PCA.*

Table 1 shows the features of the energy data used in the evaluation. For each cluster, we have analysed various feature ranges, including the number of data points for a Point of Distribution (POD) representing a prosumer energy meter, the minimum, mean, and maximum amounts of active energy recorded by a POD, and the minimum and maximum time intervals, expressed in the format yyyy/mm/dd/hh, during which the load is generated.

The hourly energy values were normalized using the min-max scaling method [61], rescaling them in the range [0,1]:

$$E_{norm} = \frac{E - E_{min}}{E_{max} - E_{min}} \tag{22}$$

where $E_{norm}$ is the normalized energy value, $E_{max}$ and $E_{min}$ are the maximum and minimum energy values within the set of energy values that correspond to the four measurements collected at 15-minute intervals during the hour.

*Table 1. The energy data set features*

|  | **Cluster 0** | **Cluster 1** | **Cluster 2** |
|---|---|---|---|
| LENGTH (pod's datapoints number) | 145912 | lower bound: 74504 upper bound: 84576 | lower bound: 134304 upper bound: 137280 |
| MIN_ACTIVE_LOAD | 0.000 | 0.000 | 0.000 |
| MAX_ACTIVE_LOAD | lower bound: 4.925 upper bound: 302.000 | lower bound: 0.551 upper bound:  4.268 | lower bound: 5.360 upper bound:  121.328 |
| MEAN_ACTIVE_LOAD | lower bound: 0.327 upper bound: 71.903 | lower bound: 0.024 upper bound:  0.779 | lower bound: 0.426 upper bound:   9.056 |
| MIN_DHH (yyyyMMddhhmm) | 201501010000 | 201610010000 | lower bound: 201504010000 upper bound: 201505010000 |
| MAX_DHH (yyyyMMddhhmm) | 201902282345 | lower bound: 201901312345 upper bound: 201902282345 | 201902282345 |
| STD_ACTIVE_LOAD | lower bound: 0.262 upper bound: 50.950 | lower bound: 0.023 upper bound: 0.537 | lower bound: 0.152 upper bound: 6.615 |

The min-max scaling to normalize the energy values is a suitable approach when dealing with renewable energy curves that do not follow a Gaussian distribution. The deviation from such distribution can be attributed to several factors, such as clear sky index, azimuth Angle, shading, solar panel placement, or meteorological conditions. The normalized inputs improve the convergence of the local model learning while mitigating the impact of outliers in data. Also, the model can generalize better for new data because the training was done on normalized features that are consistent across different scales.

A LSTM (Long Short-Term Memory) neural network is trained on each local prosumer node. The LSTM network used has two layers of LSTM cells, and each LSTM cell has a hidden size of 64, which denotes that it has 64 neurons (see Figure 4). To prevent the overfitting phenomena, a dropout layer with a rate of 0.5 is applied on each layer. In addition to LTSM layers, we have used two fully connected layers, each of them with 64 features. Between the fully connected layers, a dropout layer with a dropout rate of 0.5 is also placed. The Adam optimizer with a learning rate of 0.0003 is used to optimize the LSTM network.

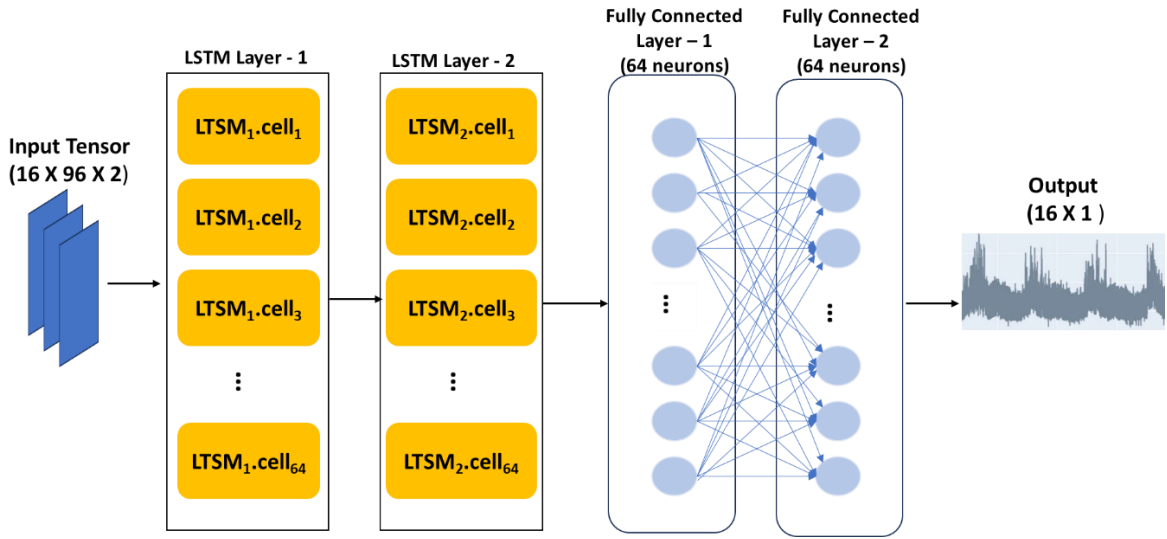

*Figure 4: The neural network architecture used for energy prediction.*

The neural network architecture was implemented in PyTorch [62]. To efficiently handle the computation of the loss function on each local prosumer node, we leverage PyTorch's chart compilation feature. This allows us to precompile and optimize a computation graph that includes all phases of the loss function, reducing wasted calculations, improving the overall efficiency, and mitigating timing-related issues of our model.

The neural network is used to predict energy values with a lead time of 4. This means that we forecast energy values one hour ahead of the current time because our dataset contains four samples at 15-minute intervals for each hour. As input, we will consider sequences with the length set to 96 energy readings, which means that for each prediction, we consider the values of the energy readings from the entire day before, encompassing 24 hours. A batch size of 16 is used during training.

The amount of energy data samples ($d_i$) used to train the LSTM neural network on each local node ($n_i$) are varying in size and are determined as:

$$d_i = \frac{|dataSet| * ratio_{n_i}}{\sum_{j=1}^{N} ratio_{n_i}} \tag{23}$$

where: $ratio_{n_i}$ is the percentage of energy data from the entire cluster available in the node $n_i$.

The communication process, at the level of each cluster, was implemented with a specific number of communication rounds between the local nodes and the central node. The process involves the central node initially sending the models to the local nodes, and then the local nodes sending back adjusted model parameters $W$.

On each cluster, we have used the defined WOA to learn the optimal global model. During WOA optimization for $\frac{T}{2}$ iterations (where T represents the total number of iterations of WOA), each local node had a 50% chance of sending its updated model parameters due to local training. This stochastic process allows for training on local data at each iteration while estimating the loss by exchanging the models among all local nodes within the same cluster. The number of communication rounds is:

$$\gamma = n^2, n = |C| \tag{24}$$

where *n* represents the number of local nodes within the cluster.

To determine the performance of the global prediction model using the best weight vector identified by the WOA algorithm, we calculated the mean square error (MSE) and the mean absolute error (MAE) for the training and testing. The MSE offers a view on the global model's ability to reduce large errors significantly and prioritize precise predictions, while the MAE shows model robustness to outliers and overall accuracy in predicting energy values. The results obtained have been compared with a popular state-of-the-art method the federated average where the model parameters are aggregated by averaging them across the participating nodes [14]. We have maintained a consistent data distribution between the two evaluations to make a fair comparison of the results.

Tables 2-4 present the values obtained for these metrics for each local nod in each cluster considering both FL with WOA and federated average.

*Table 2: FedWOA accuracy compared with FedAVG for clusters with ID 0 nodes.*

| Node ID | MSE | | | | MAE | | | |
| --- | --- | --- | --- | --- | --- | --- | --- | --- |
| | Train | | Test | | Train | | Test | |
| | FedWOA | FedAVG | FedWOA | FedAVG | FedWOA | FedAVG | FedWOA | FedAVG |
| P_000004 | **0.008284** | 0.037937 | **0.019984** | 0.070258 | **0.069696** | 0.166328 | **0.114504** | 0.247359 |
| P_000006 | 0.006901 | **0.002212** | **0.007506** | 0.010177 | 0.078411 | **0.033689** | 0.069709 | **0.069060** |
| P_000010 | 0.013317 | **0.008977** | 0.012887 | **0.008231** | 0.101358 | **0.038855** | 0.099493 | **0.035694** |
| P_000013 | **0.004946** | 0.023111 | **0.006870** | 0.029826 | **0.049691** | 0.129460 | **0.057932** | 0.148497 |
| P_000107 | **0.003999** | 0.024867 | **0.007489** | 0.039646 | **0.044039** | 0.138278 | **0.056876** | 0.173869 |
| P_0048AB | **0.003603** | 0.018498 | **0.008010** | 0.017770 | **0.045826** | 0.119876 | **0.071150** | 0.097609 |
| P_004CB7 | 0.005758 | **0.002851** | **0.006722** | 0.007378 | 0.070555 | **0.038802** | 0.072343 | **0.058888** |
| P_00506E | **0.009390** | 0.029393 | **0.009049** | 0.028525 | **0.079284** | 0.129961 | **0.075593** | 0.128199 |
| P_00509B | **0.016553** | 0.039929 | **0.023252** | 0.049697 | **0.104738** | 0.148296 | **0.116029** | 0.158162 |
| P_005169 | **0.010605** | 0.030274 | **0.003903** | 0.016410 | **0.078645** | 0.125971 | **0.026814** | 0.112718 |

The results show that in most of the evaluation cases, our proposed FL model outperforms the federated average being more effective for the problem of energy prediction of prosumers renewable generation.

*Table 3: FedWOA accuracy compared with FedAVG for the cluster with ID 1 nodes.*

| Node ID | MSE | | | | MAE | | | |
|---|---|---|---|---|---|---|---|---|
| | Train | | Test | | Train | | Test | |
| | FedWOA | FedAVG | FedWOA | FedAVG | FedWOA | FedAVG | FedWOA | FedAVG |
| P_0005FT | **0.003829** | 0.005834 | **0.008828** | 0.011027 | **0.035584** | 0.047204 | **0.056174** | 0.061671 |
| P_0008D9 | **0.002339** | 0.017111 | **0.001618** | 0.012010 | **0.025652** | 0.112864 | **0.026775** | 0.097650 |
| P_00119D | **0.003394** | 0.013606 | **0.004573** | 0.005517 | **0.029788** | 0.093690 | **0.043059** | 0.043576 |
| P_001236 | **0.003529** | 0.009005 | **0.001962** | 0.009590 | **0.032112** | 0.067071 | **0.031365** | 0.079683 |
| P_001A05 | **0.004125** | 0.006001 | **0.012879** | 0.013544 | **0.031095** | 0.049709 | **0.065968** | 0.067660 |
| P_00280E | **0.002557** | 0.006015 | **0.002921** | 0.026353 | **0.028948** | 0.058552 | **0.031604** | 0.134193 |
| P_003F4D | **0.001874** | 0.009889 | **0.003749** | 0.010357 | 0.021277 | **0.090896** | **0.035885** | 0.085141 |
| P_00410C | **0.002537** | 0.006280 | **0.002959** | 0.003047 | **0.026860** | 0.055113 | 0.038478 | **0.031183** |
| P_004613 | 0.005511 | **0.027361** | **0.004873** | 0.029442 | **0.035097** | 0.117643 | **0.044560** | 0.136422 |
| P_0056CD | **0.002485** | 0.081751 | **0.001650** | 0.008533 | 0.027262 | **0.256203** | **0.028540** | 0.075401 |
| P_005D9A | **0.001543** | 0.001642 | **0.004156** | 0.003016 | 0.027777 | **0.026606** | 0.043456 | **0.026326** |
| P_007C29 | **0.009848** | 0.048921 | **0.001579** | 0.000679 | **0.039781** | 0.189627 | 0.031253 | **0.013336** |
| P_0080F1 | **0.006916** | 0.018270 | **0.002022** | 0.017550 | **0.042904** | 0.099059 | **0.029381** | 0.116548 |

*Table 4: FedWOA accuracy compared with FedAVG for the cluster with ID 2 nodes.*

| Node ID | MSE | | | | MAE | | | |
|---|---|---|---|---|---|---|---|---|
| | Train | | Test | | Train | | Test | |
| | FedWOA | FedAVG | FedWOA | FedAVG | FedWOA | FedAVG | FedWOA | FedAVG |
| P_000001 | **0.014452** | 0.050956 | **0.035215** | 0.069809 | **0.099791** | 0.182951 | **0.153632** | 0.186752 |
| P_000410 | **0.005756** | 0.042573 | **0.012841** | 0.033146 | **0.062664** | 0.188013 | **0.093337** | 0.137788 |
| P_004A72 | **0.012650** | 0.020478 | 0.030867 | **0.001029** | **0.096708** | 0.104756 | 0.173106 | **0.022626** |
| P_00680A | **0.007578** | 0.012626 | **0.021141** | 0.044094 | **0.077020** | 0.100150 | **0.118028** | 0.165886 |
| P_006A55 | **0.006136** | 0.043906 | **0.023435** | 0.095025 | 0.057145 | **0.191260** | **0.122700** | 0.265007 |
| P_007209 | **0.004594** | 0.028107 | **0.012049** | 0.037411 | **0.052845** | 0.153329 | **0.089790** | 0.158108 |
| P_0091D7 | **0.002624** | 0.037981 | **0.007879** | 0.044147 | **0.037102** | 0.186407 | **0.068846** | 0.186897 |
| P_009FA7 | 0.015610 | **0.010084** | **0.022303** | 0.017734 | 0.108311 | **0.067072** | 0.134834 | **0.076131** |
| P_00B211 | **0.006312** | 0.041473 | 0.014426 | **0.003195** | **0.051929** | 0.190132 | 0.116661 | **0.040575** |

Table 5 shows comparatively the results on average per cluster of the FL with WOA providing more accurate energy prediction results.

*Table 5: Error table for FedWOA on cluster 2*

| Cluster ID | Solution | MSE_TRAIN | MSE_VAL | MSE_TEST | MAE_TRAIN | MAE_VAL | MAE_TEST |
|---|---|---|---|---|---|---|---|
| C0 | FedWOA | **0.008336** | **0.010612** | **0.010567** | **0.072224** | **0.079599** | **0.076044** |
| | FedAVG | 0.021805 | 0.025414 | 0.027792 | 0.106952 | 0.110631 | 0.123006 |
| C1 | FedWOA | **0.003888** | **0.005203** | **0.004136** | **0.031087** | **0.036358** | **0.038961** |
| | FedAVG | 0.019360 | 0.011760 | 0.011590 | 0.097249 | 0.067437 | 0.074522 |
| C2 | FedWOA | **0.008413** | **0.014036** | **0.020017** | **0.071503** | **0.097329** | **0.118993** |
| | FedAVG | 0.032021 | 0.038669 | 0.038399 | 0.151563 | 0.153327 | 0.137752 |

We calculated, for each cluster, the percentage of improvement that our algorithm brings compared to the classic variant using:

$$improvement = \frac{error_{avg}}{error_{WoA}} * 100 \qquad (25)$$

where the error represents the average MSE or MAE error computed for training, validation, or testing. Figure 7 presents the percentage of improvement computed for MSE, and MAE per cluster of prosumers. The results show the effectiveness of our FedWOA algorithm, compared to the state-of-the-art FedAVG. FedWOA outperforms FedAVG in terms of Mean Square Error and Mean Absolute Error metrics for the federated energy prediction processes. The improvements of FedWOA compared with FedAVG are between, 22,6% in the worst case and 27,5 % in the best case for MSE in the prediction validation phase and between 13,8% in the worst case and 18,5% in the best case for MAE for prediction validation.

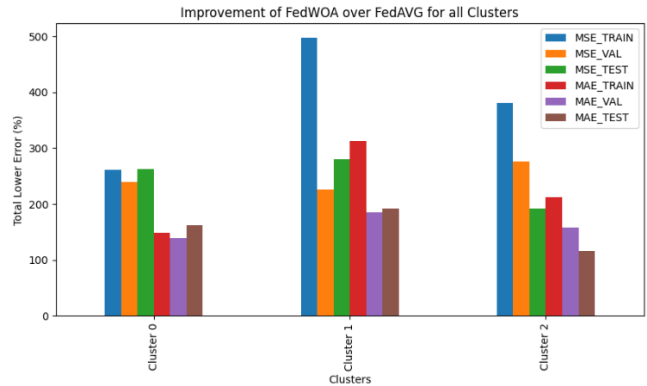

*Figure 5: Improvement of FedWOA over FedAVG*

We also compared our FL approach using WOA with the FL average considering the predicted energy curve. Figure 6 shows, for each of the three clusters, the actual energy curve to be predicted, the energy curve predicted with WOA, and the energy curve predicted with the classical FL approach. Analyzing these graphs, we can see that the energy curve predicted with WOA more accurately approximates the real energy curve than the one predicted with the classical FL approach.

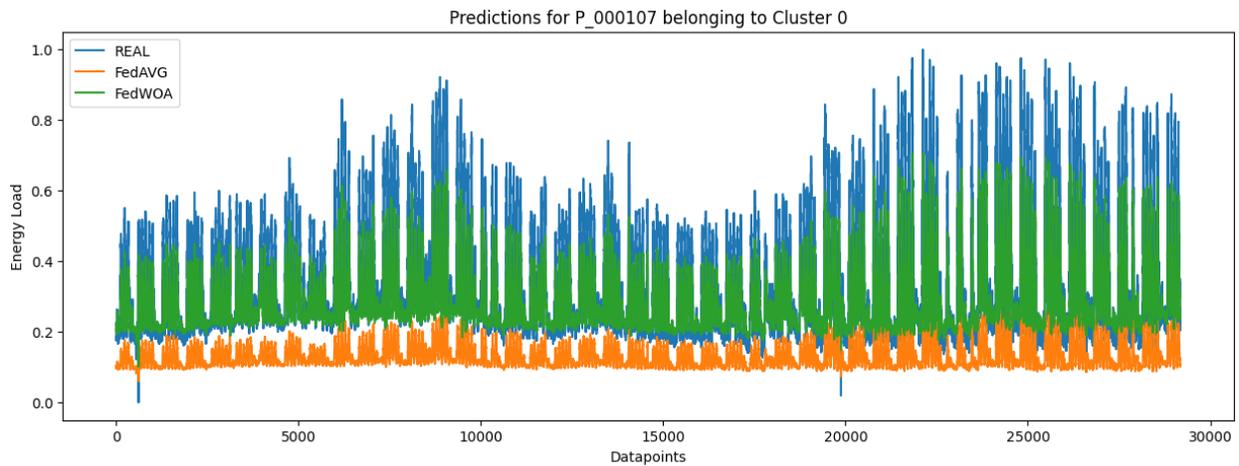

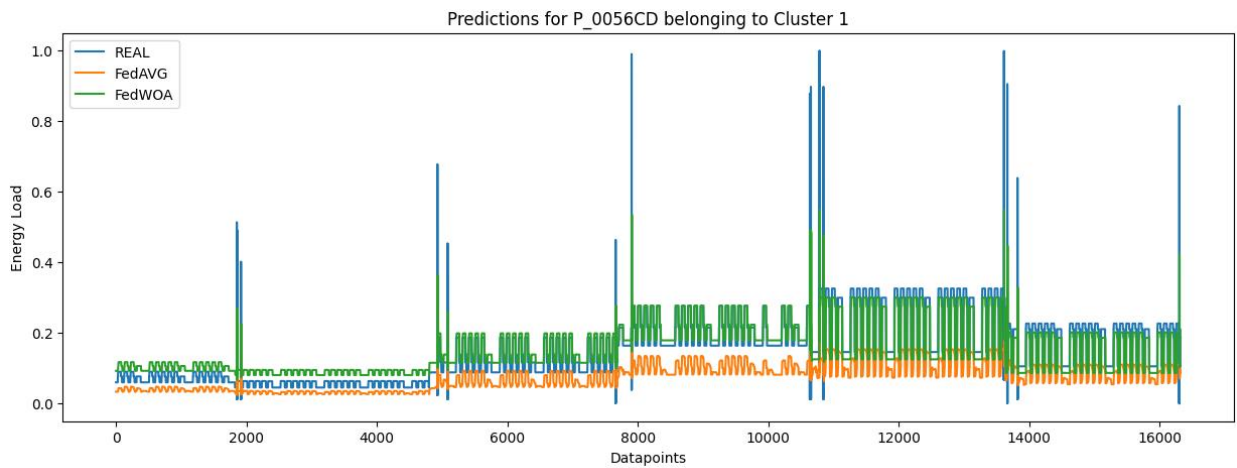

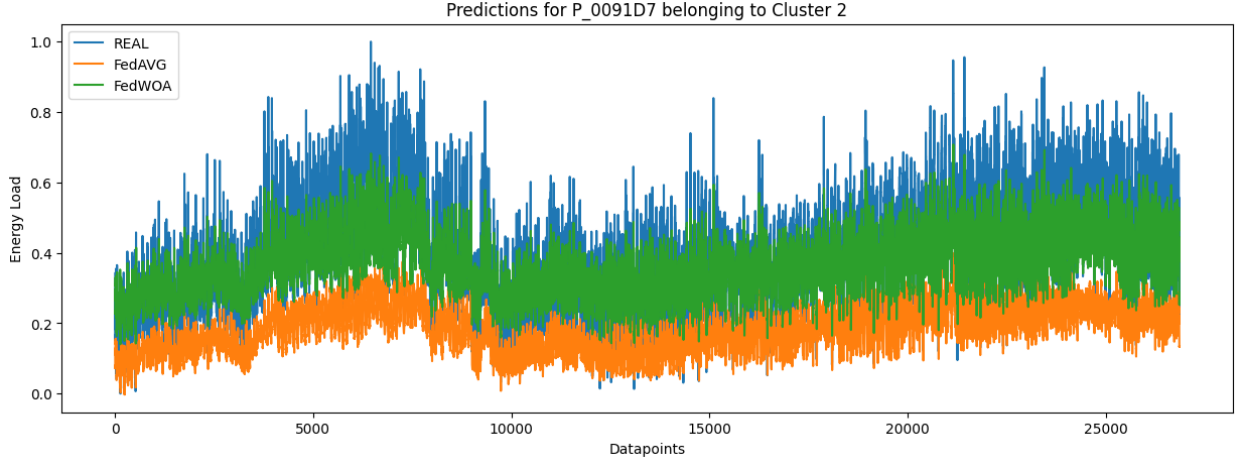

*Figure 6: Energy curve predicted for different prosumers of the clusters 0,1 and 2.*

## 5. Discussion

In this section, we discuss the convergence rate and diversity of a WOA for FL considering the prediction of the prosumers energy generation use case. Such metrics offer a good view of how fast the WOA provides a good solution for FL-based energy prediction and how diverse are the proposed solutions across different iterations. Experiments have been conducted to comparatively determine the fitness value, convergence rate, and diversity on different configurations.

Fitness value measures the quality of the solution return by WOA and is computed using the defined loss function $L$ for the local and global federated models. The convergence rate is used to quantify the improvement rate in fitness values during the algorithm iterations and it is computed as [59]:

$$rate_{convergence} = 1 - (|\frac{L_{opt} - L_t}{L_{opt} - L_{Goal}}|)^{1/T} \qquad (26)$$

where $L_{opt}$ is the best fitness obtained by our algorithm in the $T$ iterations, $L_t$ is the fitness at the current iteration $t$, $L_{Goal}$ is the target fitness that in our case has the ideal value of (i.e. no loss in the prediction process) and $T$ is the number of iterations. The diversity is used to assess how well the WOA explores the search space for determining the optimal vector of parameters for the global federated model out of the local models learned [60]. The diversity is computed for each iteration using the Euclidean distance:

$$div = \sum_{i=1}^{N} \sum_{j=1}^{M} \sqrt{(\overrightarrow{W_i} - \overrightarrow{W_j})^2} \qquad (27)$$

where $\overrightarrow{W_i}$ and $\overrightarrow{W_j}$ are the positions of the individuals (i.e., weights vectors) in the search space.

Figure 7 shows the evolution of average loss (i.e., the fitness value of the best individual) during FedWOA iterations for each of the three clusters. Analysing these plots, we notice that, for each of the three clusters FedWOA can find, after ten iterations, a solution with a fitness value very close to zero thus it efficiently explores the search space to identify a solution that is very close to the global optimum. However, in the case of the second cluster, we notice that although in the first iteration, the loss validation function decreases, starting with the second iteration, it increases and remains constant until interaction with the fourth one, when it begins to decrease again. This is due to the way the dataset was divided, namely that the validation dataset has less noise than the training dataset.

We also made a comparison in terms of the average training loss and the average validation loss of the FedWOA compared with FedAVG. The results show the evolution of average training loss and average validation loss from one epoch to another for each of the three clusters in the case of the FedAVG. Analysing these plots, we notice that our FedWOA can reach a better loss (i.e., a better fitness) in only ten iterations as opposed to the 50 epochs ran in the case of FedAVG. Also, it can be observed that in the FedAVG approach, the loss slightly reduces and continues to fluctuate from one epoch to another, while in the case of FedWOA, the loss reduces significantly and continues to decrease from one iteration to another.

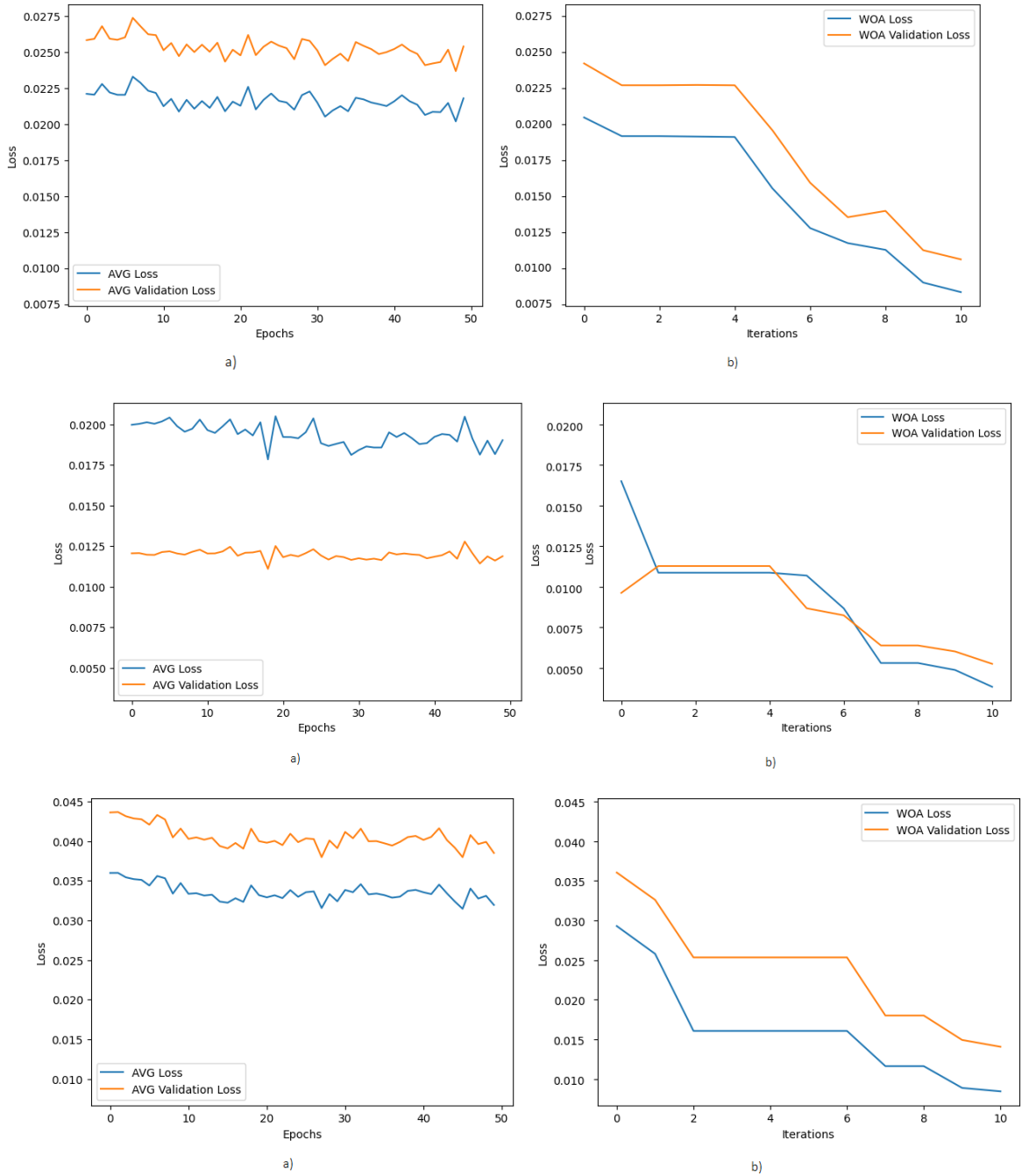

*Figure 7: The fitness variation for each cluster of prosumers*

Figure 8 represents the average convergence rate chart for the FedWOA solution for each of the three clusters. We observe an improvement in fitness value from one iteration interval to the next. Also, there are no fluctuations in the convergence rate, but only intervals in which it stagnates and then increases again and approaches the optimal value. This indicates that the algorithm is converging fast towards the best solution. The plateau recorded in the graphs means that the best individual can remain the same for several iterations until another better individual is identified (i.e., an individual with a lower loss value).

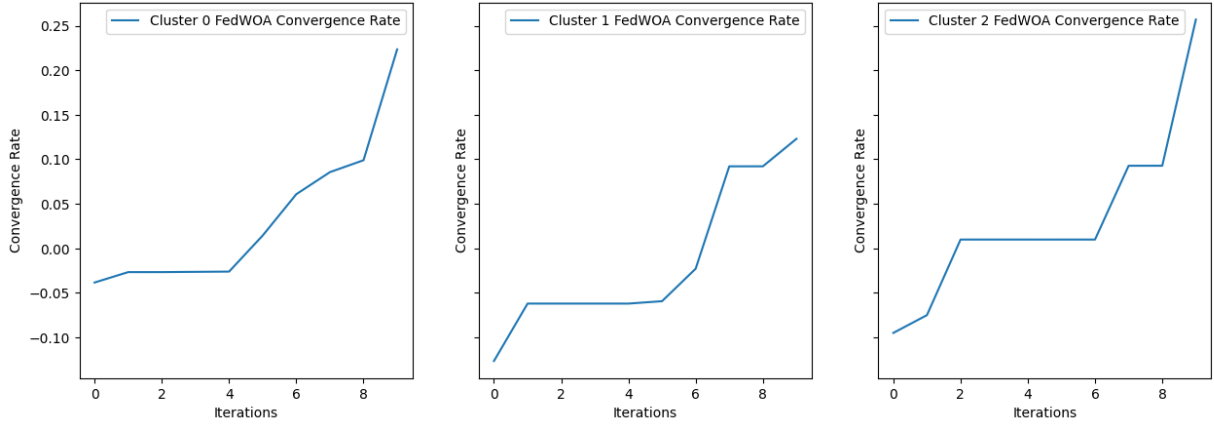

*Figure 8: Convergence rate of FedWOA for each of the three clusters*

The diversity measurements for each cluster are presented in Figure 9. What is observed from all three graphs corresponding to the three clusters in which the WOA is applied, is that there are stages where the algorithm's diversity decreases indicating exploitation of the current search area, or stages where the algorithm's diversity increases, indicating exploration of new areas. More exactly, we observe that in the first iterations, the algorithm mainly performs an exploration of the search space, and therefore the diversity increases, while in the last iterations, the algorithm focuses on exploitation of the search space to converge towards the best solution.

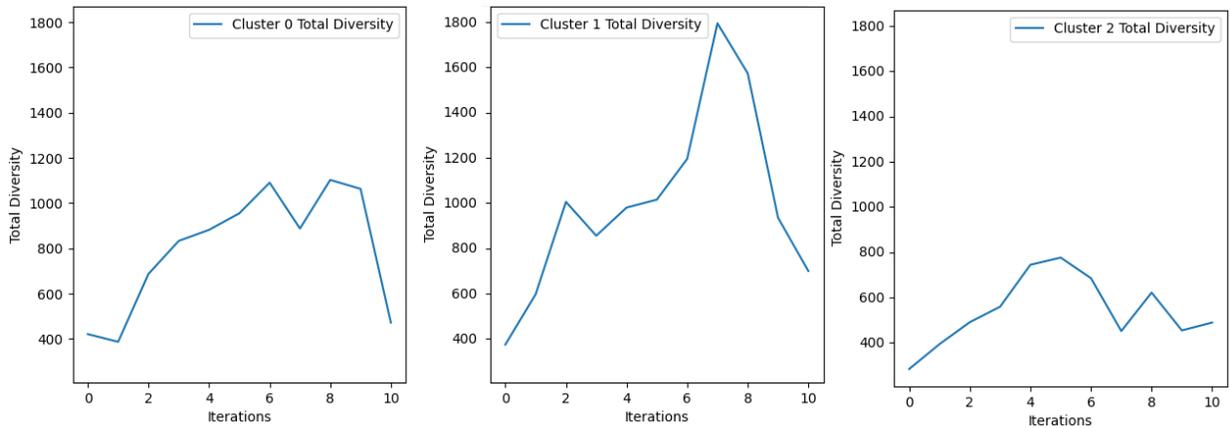

*Figure 9: Diversity of FedWOA for each of the three clusters*

## 6. Conclusions

In this paper, we proposed a federated machine learning model for predicting renewable energy production based on time series energy data from local prosumer nodes. The novelty of our approach is the utilization of the WOA to aggregate global prediction models from the weights of local LTSM neural network models to address challenges related to prosumers data heterogeneity, and variations in generation patterns. The FedWOA solution identifies the near-optimal vector of weights to construct the global shared model and then is subsequently transmitted to the local nodes to improve the prediction quality at the prosumer site while the issue of non-IID data was addressed by using K-means clustering to group the prosumers with similar scale of energy data.

FedWOA has demonstrated efficacy in improving the accuracy of energy prediction models. Compared to FedAVG, our solution provides better energy prediction accuracy with an average improvement of about 25% for MSE and 16% for MAE in the prediction validation phase, showing good convergence and reduced loss. Our findings reveal that using decentralized time series energy data sources, and collaborative global model optimization using WOA, we can achieve more precise forecasts for small-scale energy prosumers.

While our paper has provided valuable insights on federated solutions for energy predictions, there remain opportunities for further exploration, particularly in improving the efficiency of our model in distributed smart grid environments while considering diverse energy generation sources. To enhance scalability, we plan to consider new and more efficient models for loss computation to address the potential computational bottleneck in the case of model training on many edge nodes.

## Acknowledgement

This work has been conducted within the DEDALUS project, grant number 101103998, funded by the European Commission as part of the Horizon Europe Framework Programme.

## References

1. Jaap Wieringa, P.K. Kannan, Xiao Ma, Thomas Reutterer, Hans Risselada, Bernd Skiera, Data analytics in a privacy-concerned world, Journal of Business Research, Volume 122, 2021, Pages 915-925, ISSN 0148-2963.
2. L. Toderean, V. R. Chifu, T. Cioara, I. Anghel and C. B. Pop, "Cooperative Games Over Blockchain and Smart Contracts for Self-Sufficient Energy Communities," in IEEE Access, vol. 11, pp. 73982-73999, 2023, doi: 10.1109/ACCESS.2023.3296258.
3. Mehdi Mehdinejad, Heidarali Shayanfar, Behnam Mohammadi-Ivatloo, Peer-to-peer decentralized energy trading framework for retailers and prosumers, Applied Energy, Volume 308, 2022, 118310, ISSN 0306-2619,.
4. Pop, C.; Cioara, T.; Antal, M.; Anghel, I.; Salomie, I.; Bertoncini, M. Blockchain Based Decentralized Management of Demand Response Programs in Smart Energy Grids. Sensors 2018, 18, 162.
5. Antal, M.; Toderean, L.; Cioara, T.; Anghel, I. Hybrid Deep Neural Network Model for Multi-Step Energy Prediction of Prosumers. Appl. Sci. 2022, 12, 5346. https://doi.org/10.3390/app12115346
6. Mel Keytingan M. Shapi, Nor Azuana Ramli, Lilik J. Awalin, Energy consumption prediction by using machine learning for smart building: Case study in Malaysia, Developments in the Built Environment, Volume 5, 2021, 100037, ISSN 2666-1659, https://doi.org/10.1016/j.dibe.2020.100037.
7. Runge, J.; Zmeureanu, R. A Review of Deep Learning Techniques for Forecasting Energy Use in Buildings. Energies 2021, 14, 608. https://doi.org/10.3390/en14030608
8. Adewole, K.S., Torra, V. DFTMicroagg: a dual-level anonymization algorithm for smart grid data. Int. J. Inf. Secur. 21, 1299–1321 (2022). https://doi.org/10.1007/s10207-022-00612-8
9. Tianshi Mu, Yuyang Lai, Guocong Feng, Huahui Lyu, Hang Yang, Jianfeng Deng, A user-friendly attribute-based data access control scheme for smart grids, Alexandria Engineering Journal, Volume 67, 2023, Pages 209-217, ISSN 1110-0168, https://doi.org/10.1016/j.aej.2022.12.041.
10. Mitrea, D.; Cioara, T.; Anghel, I. Privacy-Preserving Computation for Peer-to-Peer Energy Trading on a Public Blockchain. Sensors 2023, 23, 4640. https://doi.org/10.3390/s23104640
11. Martinez, J., Ruiz, A., Puelles, J., Arechalde, I., Miadzvetskaya, Y. (2020). Smart Grid Challenges Through the Lens of the European General Data Protection Regulation. In: Siarheyeva, A., Barry, C., Lang, M., Linger, H., Schneider, C. (eds) Advances in Information Systems Development. ISD 2019. Lecture Notes in Information Systems and Organisation, vol 39. Springer, Cham. https://doi.org/10.1007/978-3-030-49644-9_7
12. Antal, M.; Mihailescu, V.; Cioara, T.; Anghel, I. Blockchain-Based Distributed Federated Learning in Smart Grid. Mathematics 2022, 10, 4499. https://doi.org/10.3390/math10234499
13. Liu, Haizhou, et al. "A federated learning framework for smart grids: Securing power traces in collaborative learning." *arXiv preprint arXiv:2103.11870* (2021).
14. McMahan, Brendan, et al. "Communication-efficient learning of deep networks from decentralized data." Artificial intelligence and statistics. PMLR, 2017.
15. He, Chaoyang, et al. "Fedml: A research library and benchmark for federated machine learning." arXiv preprint arXiv:2007.13518 (2020).
16. T. Li, A. K. Sahu, A. Talwalkar and V. Smith, "Federated Learning: Challenges, Methods, and Future Directions," in IEEE Signal Processing Magazine, vol. 37, no. 3, pp. 50-60, May 2020, doi: 10.1109/MSP.2020.2975749.
17. L. U. Khan, W. Saad, Z. Han, E. Hossain and C. S. Hong, "Federated Learning for Internet of Things: Recent Advances, Taxonomy, and Open Challenges," in IEEE Communications Surveys & Tutorials, vol. 23, no. 3, pp. 1759-1799, thirdquarter 2021, doi: 10.1109/COMST.2021.3090430.

18. Cristina Bianca Pop, Tudor Cioara, Ionut Anghel, Marcel Antal, Viorica Rozina Chifu, Claudia Antal, Ioan Salomie, Review of bio-inspired optimization applications in renewable-powered smart grids: Emerging population-based metaheuristics, Energy Reports, Volume 8, 2022, Pages 11769-11798, ISSN 2352-4847,.
19. Seyedali Mirjalili, Andrew Lewis, The Whale Optimization Algorithm, Advances in Engineering Software, Volume 95, 2016, Pages 51-67, ISSN 0965-9978, ttps://doi.org/10.1016/j.advengsoft.2016.01.008.
20. Brodzicki, A.; Piekarski, M.; Jaworek-Korjakowska, J. The Whale Optimization Algorithm Approach for Deep Neural Networks. Sensors 2021, 21, 8003. https://doi.org/10.3390/s21238003
21. Rana, N., Latiff, M.S.A., Abdulhamid, S.M. et al. Whale optimization algorithm: a systematic review of contemporary applications, modifications and developments. Neural Comput & Applic 32, 16245–16277 (2020).
22. S. Na, L. Xumin and G. Yong, "Research on k-means Clustering Algorithm: An Improved k-means Clustering Algorithm," 2010 Third International Symposium on Intelligent Information Technology and Security Informatics, Jian, China, 2010, pp. 63-67, doi: 10.1109/IITSI.2010.74.
23. M. N. Fekri, K. Grolinger, and S. Mir, Distributed load forecasting using smart meter data: Federated learning with recurrent neural networks, International Journal of Electrical Power & Energy Systems, vol. 137, 2022.
24. A. Moradzadeh, H. Moayyed, B. Mohammadi-Ivatloo, A. P. Aguiar, and A. Anvari- Moghaddam, "A secure federated deep learning-based approach for heating load demand forecasting in building environment," IEEE Access, vol. 10, pp. 5037–5050, 2021.
25. M. Savi and F. Olivadese, "Short-term energy consumption forecasting at the edge: A federated learning approach," IEEE Access, vol. 9, pp. 95 949–95 969, 2021.
26. J. D. Fernandez, S. P. Menci, C. M. Lee, A. Rieger, and G. Fridgen, "Privacy preserving federated learning for residential short-term load forecasting," Applied Energy, vol. 326, p. 119 915, 2022.
27. G. Zhang, S. Zhu, and X. Bai, "Federated learning-based multi-energy load forecasting method using cnn-attention-lstm model," Sustainability, vol. 14, no. 19, p. 12 843, 2022.
28. Shi, Y.; Xu, X. Deep Federated Adaptation: An Adaptative Residential Load Forecasting Approach with Federated Learning. *Sensors* 2022, *22*, 3264.
29. Gholizadeh, N.; Musilek, P. Federated learning with hyperparameter-based clustering for electrical load forecasting. Internet Things 2022, 17, 100470.
30. Yi Wang, Ning Gao, and Gabriela Hug, Personalized Federated Learning for Individual Consumer Load Forecasting, CSEE JOURNAL OF POWER AND ENERGY SYSTEMS, VOL. 9, NO. 1, JANUARY 202
31. Yixing Liu, Zhen Dong, Bo Liu, Yiqiao Xu, Zhengtao Ding, FedForecast: A federated learning framework for short-term probabilistic individual load forecasting in smart grid, International Journal of Electrical Power & Energy Systems, Volume 152, 2023, 109172, ISSN 0142-0615
32. V. Venkataramanan, S. Kaza and A. M. Annaswamy, "DER Forecast Using Privacy-Preserving Federated Learning," in *IEEE Internet of Things Journal*, vol. 10, no. 3, pp. 2046-2055, 1 Feb.1, 2023, doi: 10.1109/JIOT.2022.3157299.
33. L. Ding, J. Wu, C. Li, A. Jolfaei and X. Zheng, "SCA-LFD: Side-Channel Analysis-Based Load Forecasting Disturbance in the Energy Internet," in *IEEE Transactions on Industrial Electronics*, vol. 70, no. 3, pp. 3199-3208, March 2023, doi: 10.1109/TIE.2022.3170641.
34. Y. Guo, D. Wang, A. Vishwanath, C. Xu, and Q. Li, "Towards federated learning for HVAC analytics: A measurement study," in Proc. 11th ACM Int. Conf. Future Energy Syst., 2020, pp. 68–73
35. Atharvan Dogra, Ashima Anand, Jatin Bedi, Consumers profiling based federated learning approach for energy load forecasting, Sustainable Cities and Society, Volume 98, 2023, 104815, ISSN 2210-6707
36. Brisimi, T.S.; Chen, R.; Mela, T.; Olshevsky, A.; Paschalidis, I.C.; Shi, W. Federated learning of predictive models from federated electronic health records. Int. J. Med. Inform. 2018, 112, 59–67. [CrossRef]
37. Fang, L.; Liu, X.; Su, X.; Ye, J.; Dobson, S.; Hui, P.; Tarkoma, S. Bayesian inference federated learning for heart rate prediction. In Proceedings of the International Conference on Wireless Mobile Communication and Healthcare, Virtual Event, 19 November 2020; Springer: Cham, Switzerland, 2020; pp. 116–130. 105.
38. Brophy, E.; De Vos, M.; Boylan, G.; Ward, T. Estimation of continuous blood pressure from ppg via a federated learning approach. Sensors 2021, 21, 6311.
39. Lo, J.; Timothy, T.Y.; Ma, D.; Zang, P.; Owen, J.P.; Zhang, Q.; Wang, R.K.; Beg, M.F.; Lee, A.Y.; Sarunic, M.V.; et al. Federated learning for microvasculature segmentation and diabetic retinopathy classification of OCT data. Ophthalmol. Sci. 2021, 1, 100069.
40. Zhang, L.; Shen, B.; Barnawi, A.; Xi, S.; Kumar, N.; Wu, Y. FedDPGAN: Federated differentially private generative adversarial networks framework for the detection of COVID-19 pneumonia. Inf. Syst. Front. 2021.

41. Dou, Q.; So, T.Y.; Jiang, M.; Liu, Q.; Vardhanabhuti, V.; Kaissis, G.; Li, Z.; Si, W.; Lee, H.H.C.; Yu, K.; et al. Federated deep learning for detecting COVID-19 lung abnormalities in CT: A privacy-preserving multinational validation study. NPJ Digit. Med. 2021, 4, 60.
42. Abdul Salam, M.; Taha, S.; Ramadan, M. COVID-19 detection using federated machine learning. PLoS ONE 2021, 16, e0252573.
43. Vaid, A.; Jaladanki, S.K.; Xu, J.; Teng, S.; Kumar, A.; Lee, S.; Somani, S.; Paranjpe, I.; De Freitas, J.K.; Wanyan, T.; et al. Federated learning of electronic health records to improve mortality prediction in hospitalized patients with COVID-19: Machine learning approach. JMIR Med. Inform. 2021, 9, e24207.
44. Chen, Y.; Qin, X.; Wang, J.; Yu, C.; Gao, W. FedHealth: A federated transfer learning framework for wearable healthcare. IEEE Intell. Syst. 2020, 35, 83–93.
45. E. Arnold, O. Y. Al-Jarrah, M. Dianati, S. Fallah, D. Oxtoby, and A. Mouzakitis, "A survey on 3d object detection methods for autonomous driving applications," IEEE Transactions on Intelligent Transportation Systems, vol. 20, no. 10, pp. 3782–3795, 2019.
46. Qi, Y.; Hossain, M.S.; Nie, J.; Li, X. Privacy-preserving blockchain-based federated learning for traffic flow prediction. Futur. Gener. Comput. Syst. 2021, 117, 328–337.
47. Mazin Abed Mohammed, Abdullah Lakhan, Karrar Hameed Abdulkareem, Dilovan Asaad Zebari, Jan Nedoma, Radek Martinek, Seifedine Kadry, Begonya Garcia-Zapirain, Homomorphic federated learning schemes enabled pedestrian and vehicle detection system, Internet of Things, Volume 23, 2023, 100903, ISSN 2542-6605,.
48. Alohali, M.A.; Aljebreen, M.; Nemri, N.; Allafi, R.; Duhayyim, M.A.; Ibrahim Alsaid, M.; Alneil, A.A.; Osman, A.E. Anomaly Detection in Pedestrian Walkways for Intelligent Transportation System Using Federated Learning and Harris Hawks Optimizer on Remote Sensing Images. *Remote Sens.* **2023**, *15*, 3092.
49. Kan Xie, Zhe Zhang, Bo Li, Jiawen Kang, Dusit Niyato, Efficient Federated Learning With Spike Neural Networks for Traffic Sign Recognition, IEEE TRANSACTIONS ON VEHICULAR TECHNOLOGY, VOL. 71, NO. 9, SEPTEMBER 2022
50. Christian Koetsier, Jelena Fiosina, Jan N. Gremmel, Jörg P. Müller, David M. Woisetschläger, Monika Sester, Detection of anomalous vehicle trajectories using federated learning, ISPRS Open Journal of Photogrammetry and Remote Sensing, Volume 4, 2022, 100013, ISSN 2667-3932, https://doi.org/10.1016/j.ophoto.2022.100013.
51. Wu, Q.; He, K.; Chen, X. Personalized Federated Learning for Intelligent IoT Applications: A Cloud-Edge Based Framework. IEEE Open J. Comput. Soc. 2020, 1, 35–44.
52. Fu, X.; Peng, R.; Yuan, W.; Ding, T.; Zhang, Z.; Yu, P.; Kadoch, M. Federated Learning-Based Resource Management with Blockchain Trust Assurance in Smart IoT. *Electronics* **2023**, *12*, 1034.
53. Park, S.; Suh, Y.; Lee, J. FedPSO: Federated Learning Using Particle Swarm Optimization to Reduce Communication Costs. *Sensors* **2021**, *21*, 600. https://doi.org/10.3390/s21020600
54. Lazzarini, R.; Tianfield, H.; Charissis, V. Federated Learning for IoT Intrusion Detection. AI 2023, 4, 509–530. https://doi.org/10.3390/ ai4030028
55. Thavavel Vaiyapuri, Shabbab Algamdi, Rajan John, Zohra Sbai, Munira Al-Helal, Ahmed alkhayyat, Deepak Gupta, Metaheuristics with federated learning enabled intrusion detection system in Internet of Things environment, Expert Systems wiley, 2022, DOI: 10.1111/exsy.13138
56. Samarakoon, S.; Bennis, M.; Saad, W.; Debbah, M. Distributed Federated Learning for Ultra-Reliable Low-Latency Vehicular Communications. IEEE Trans. Commun. 2020, 68, 1146–1159.
57. Yang, H.H.; Liu, Z.; Quek, T.Q.S.; Poor, H.V. Scheduling Policies for Federated Learning in Wireless Networks. IEEE Trans. Commun. 2020, 68, 317–333. [CrossRef]
58. Claudia Antal, Tudor Cioara, Marcel Antal, Vlad Mihailescu, Dan Mitrea, Ionut Anghel, Ioan Salomie, Giuseppe Raveduto, Massimo Bertoncini, Vincenzo Croce, Tommaso Bragatto, Federico Carere, Francesco Bellesini, Blockchain based decentralized local energy flexibility market, Energy Reports, Volume 7, 2021, Pages 5269-5288, ISSN 2352-4847, https://doi.org/10.1016/j.egyr.2021.08.118.
59. He, Jun, Lin, Guangming, Average Convergence Rate of Evolutionary Algorithms, IEEE Transactions on Evolutionary Computation,Volume 20, Issue 2, Pages 316 - 321April 2016 Article number 7122298
60. O. Olorunda and A. P. Engelbrecht, "Measuring exploration/exploitation in particle swarms using swarm diversity," in Proc. IEEE Congr. Evol. Comput., Jun. 2008, pp. 1128–1134
61. Ahsan, M.M.; Mahmud, M.A.P.; Saha, P.K.; Gupta, K.D.; Siddique, Z. Effect of Data Scaling Methods on Machine Learning Algorithms and Model Performance. Technologies 2021, 9, 52.
62. Adam Paszke, et al. 2019. PyTorch: an imperative style, high-performance deep learning library. Proceedings of the 33rd International Conference on Neural Information Processing Systems. Curran Associates Inc., Red Hook, NY, USA, Article 721, 8026–8037.